\documentclass{article}

\usepackage[final]{nips_2016}

\usepackage[utf8]{inputenc} 
\usepackage[T1]{fontenc}    
\usepackage[bookmarks=false]{hyperref}       
\usepackage{url}            
\usepackage{booktabs}       
\usepackage{amsfonts}       
\usepackage{nicefrac}       
\usepackage{microtype}      
\usepackage{amsmath}        
\usepackage{graphicx}       
\usepackage{placeins}       
\usepackage{fixltx2e}       
\usepackage{mathtools}

\DeclareMathOperator*{\argmin}{argmin}

\title{MDL-motivated compression of GLM ensembles increases interpretability and retains predictive power}

\author{Author One\qquad Author Two\\Affiliation}

\author{
  Boris Hayete\qquad Matthew Valko \qquad Alex Greenfield \qquad Raymond Yan\\
  GNS Healthcare\\
  196 Broadway, Cambridge, MA 02139 \\
  \texttt{boris@gnshealthcare.com}
}

\begin{document}

\maketitle

\begin{abstract}
  Over the years, ensemble methods have become a staple of machine learning.  Similarly, generalized linear models (\emph{GLMs}) have become very popular for a wide variety of statistical inference tasks.  The former have been shown to enhance out-of-sample predictive power and the latter possess easy interpretability.  Recently, ensembles of GLMs have been proposed as a possibility.  On the downside, this approach loses the interpretability that GLMs possess.  We show that minimum description length (\emph{MDL})-motivated compression of the inferred ensembles can be used to recover interpretability without much, if any, downside to performance and illustrate on a number of standard classification data sets.
\end{abstract}

\section{Introduction}

Ensemble classifiers have become very popular for classification and regression tasks.  They offer the potential advantages of robustness via bootstrapping, feature prioritization, and good out-of-sample performance characteristics (\citet{chitraa}).  However, they suffer from lack of interpretability, and oftentimes features are reported as "word bags" - e.g. by feature importance (\citet{liaw}).  Generalized linear models, a venerable statistical toolchest, offer good predictive performance across a range of prediction and classification tasks, well-understood theory (advantages and modes of failure) and implementation considerations and, most importantly, excellent interpretability.  Until recently, there has been little progress in bringing together ensemble learning and GLMs, but some recent work in this area (e.g. \citet{song}) has resulted in publicly-available implementations of GLM ensembles. Nevertheless, the resulting ensembles of GLMs remain difficult to interpret.  Meantime, human understanding of models is pivotal in some fields - e.g. in translational medicine, where machine learning influences drug positioning, clinical trial design, treatment guidelines, and other outcomes that directly influence people's lives.  Improvement in performance without interpretability can be useless in such context.   To improve performance of maximum-likelihood models, \citet{carvalho} proposed to learn multiple centroids of parameter space.  Built bottom-up, such ensembles would have only a limited number of models, keeping the ensemble interpretable.  In this paper, we work from a model ensemble down.  We demonstrate that minimum description length-motivated ensemble summarization can dramatically improve interpretability of model ensembles with little if any loss of predictive power, and outline some key directions in which these approaches may evolve in the future.

\section{Methods}
\subsection{Theoretical considerations}
The problem of ML estimators being drawn to dominant solutions is well understood.  Likewise, an ensemble consensus can be drawn to the (possibly infeasible) mode, despite potentially capturing the relevant variability in the parameter space.  Relevant observations on this issue are made in \citet{carvalho}, who have proposed centroid estimators as a solution.  Working from the ensemble backwards, we use this idea as the inspiration to compress ensembles to their constituent centroids.

In order to frame the problem of ensemble summarization as that of MDL-driven compression, we consider which requirements a GLM ensemble must meet in order to be compressible, and what is required of the compression technique.  To wit, these are:
\begin{enumerate}
\item Representation
\begin{itemize}
\item The ensemble members needs to be representible as vectors in a Cartesian space
\item The ensemble needs to be "large enough" with respect to its feature set
\item The ensemble needs to have a very non-uniform distribution over features
\end{itemize}
\item Compression: the compression technique needs to
\begin{itemize}
\item capture ensemble as a number of overlapping or non-overlapping clusters 
\item provide a loss measure
\item formulate a "description length" measure
\end{itemize}
\end{enumerate}

It is easy to see that GLM ensembles can satisfy the representation requirement very directly.  It is sufficient to view ensembles of \emph{regularized} GLMs as low-dimensional vectors in a high-dimensional space.  The dimensionality of the overall space will somewhat depend on the cardinality of the ensemble, on the strictness of regularization used, on the amount of signal in the data, on the order of interactions investigated, and on other factors influencing the search space of the optimizer generating the ensemble of GLMs.  Coordinates in this space can be alternately captured by (ideally standardized) coefficients or, perhaps more meaningfully, by some function of statistical significance of the terms.  In this work, we apply the latter.

For representation, we choose a basis vector of subnetworks.  In order to identify this basis vector, we have experimented with Gaussian mixture decomposition (GMM) (finding clusters of vectors in model space) and hierarchical clustering.  For performance reasons, we present results using the latter technique, despite its shortcomings: instability and inability to fit overlapping clusters (this may lead to overfitting).  Nevertheless, in practice we find that this latter technique performs reasonably well.  Optionally, to summarize the clusters, centroids can be fit \emph{de novo} once these groups of models are identified, or medoids can be used, obviating the need for further fitting.  Here we use the first method, refitting centroids from training data on just the terms occurring in the models in a given cluster.

Lastly, Bayesian Information Criterion (\emph{BIC}) satisfies the representation scoring requirement.  The likelihood term serves as the loss function and the penalty term captures "description length" (\citet{hansen}).  

\subsection{Implementation details}
The BIC-regularized GLM ensembles were fit for binary-outcome datasets used in \citet{song} and using the software from the same paper (number of bags == 100, other settings left at defaults).  The result of this step was an ensemble $B$ which, ignoring the outcome variable and the intercepts, could be captured via a non-sparse matrix as follows:
\[
\begin{bmatrix}
    \beta_{11} & \beta_{12} & \beta_{13} & \dots  & \beta_{1n} \\
    \beta_{21} & \beta_{22} & \beta_{23} & \dots  & \beta_{2n} \\
    \vdots & \vdots & \vdots & \ddots & \vdots \\
    \beta_{d1} & \beta_{d2} & \beta_{d3} & \dots  & \beta_{dn}
\end{bmatrix}
\] 
where $d$, the ensemble dimensionality, refers to the number of fitted models and $n$ to the number of terms found in the whole fitted ensemble.  Importantly, $d$ is always an arbitrary parameter - the fact that partially motivated our study.

For each dataset, the fitted ensembles were then compressed using the following procedure.

First of all, for each ensemble we created the significance matrix S:
\[
\begin{bmatrix}
    s_{11} & s_{12} & s_{13} & \dots  & s_{1n} \\
    s_{21} & s_{22} & s_{23} & \dots  & s_{2n} \\
    \vdots & \vdots & \vdots & \ddots & \vdots \\
    s_{d1} & s_{d2} & s_{d3} & \dots  & s_{dn}
\end{bmatrix}
\] 

where $s_{ij} = -log_{10}(\mbox{p-value}(\beta_{ij}))$, and the p-value is determined from the fit of the linear model $i$ of the GLM ensemble (S is the heatmap in Figure \ref{Figure1}).  Each row of $S$ projects every model $i$ into a multivariate Cartesian space where each axis corresponds to model terms observed in the whole ensemble and each coordinate corresponds to log-scaled term significance.  Log-scaling is introduced to induce separability of models that share terms in the presence or absence of collinear covariates that would be expected to influence, but not obviate, shared terms' significance. Note that, in this representation, if $\beta_{ij}$ is missing, $s_{ij} = -log10(1) = 0$.

As an example of what the $B$ and $S$ matrices would look like after this step, for the \textbf{compressed} ensemble in Figure \ref{Figure2}, $B$ =
\[
\begin{bmatrix}
    0.081 & 0 & 0 \\
    0 & 0.4358 & 0.1917 
\end{bmatrix}
\] 
and $S$ =
\[
\begin{bmatrix}
    -log10(0.0027) & 0 & 0 \\
    0 & -log10(0.9990) & -log10(0.9999) 
\end{bmatrix}
\] 

Of course, in practice it is the full ensemble we are interested in representing in this manner en route to compressing it, so $B$ and $S$ will have their dimensions, $d$ and $n$, of $O(100) \times O(100)$ for a computational-biological application using a regularized GLM ensemble construction approach.  

Having constructed the matrix S for each ensemble in this manner, we clustered its rows (the models' coefficients) using Ward's clustering criterion (\citet{ward}) and Euclidean distance metric.   We then traversed the resulting dendrogram from left to right, using each cutpoint to perform model assignment to clusters implied by the leaves of the resulting dendrogram (Figure \ref{Figure1}).  We captured the assignment of models to k clusters produced by each step in this traversal as a column vector $m_{k}$ expressed as a categorical variable with $k$ unique values, such that $\left\vert{m_{k}}\right\vert = d$.

Then, for each model ensemble term $j$, we extracted the vector $S_{j}$ - a column slice through the matrix S for term $j$.  We next defined a linear model $$M_j := S_j = \beta_{0} + \beta_{m_k} * m_{k}$$ with $k+1$ parameters.  The success of ensemble compression via the $k$ clusters implied by $m_{k}$ could then be assessed by defining $$Cost_{k} = \dfrac{\sum_{j \in terms}BIC(M_j)}{\left\vert{terms}\right\vert} = \dfrac{\sum_{j \in terms}-2*L(M_j) + (k+1)*log(d)}{\left\vert{terms}\right\vert}$$$L$ being the log-likelihood of the model.    This evaluation of cost across clustering levels to find maximal average likelihood compression, $\argmin_{k}(Cost_{k})$, could be viewed as using the Bayes factor as the loss function for optimization, and the process of describing the model ensemble by centroids (or medoids) of the clusters of models described by $m_{k}$ could be described as an MDL-driven compression of the ensemble, using the BIC-penalized likelihood as the measure of optimal compression.

It is worth adding that the aforementioned GMM approach for cluster membership assignment, which can also be driven by BIC(\citet{fraley0}, \citet{fraley}), does not imply a specific nested membership of models in clusters, but generally results in cluster membership strongly correlated with that identified via the hierarchical model clustering technique.  While the GMM approach is more robust (e.g., it's not path-dependent, and doesn't require specification of a linkage function), it scales worse when number of terms in the ensemble is large.

\section{Results}

Using the datasets described above, we performed 3-fold cross-validation repeated three times, and for each training fold and repeat fitted medoid- and centroid-compressed model ensembles.  For each held-out fold, we then computed out-of-sample AUC for every method (Table \ref{table:Table1}).  Additionally, we performed paired one-tailed t-tests comparing medoid and centroid compression strategies to uncompressed ensembles across folds.  All AUCs arising from repeats and folds were averaged prior to t-test to avoid pseudo-replication issues.  While medoids performed slightly worse, on average, than uncompressed ensembles, centroids' performance was degraded only in the statistically "suggestive" sense (0.05 < p < 0.1).  Note that in our experience using this technique on real-world datasets, larger datasets and continuous outcomes result in even smaller, if any, degradation of performance, performance being especially degraded for logistic regression.  In other words, we believe that these results, reported for binary outcomes, are essentially a lower bound.

\section{Conclusions}
Maximum-likelihood methods identify the most likely fit in the parameter space.  However, unless the most likely fit is vastly superior to all others and is sharply defined, a rare scenario in practice, the total probability of this fit in the infinite model ensemble may be very small.  For that reason, model ensembles are thought to be superior to individual models.  Ensemble construction gains power by sampling multiple models from the parameter space but, by so doing, loses interpretability by introducing alternative parameter configurations and values.  We build on the understanding that model parameter space centroids should be sufficient to capture predictive power of large ensembles (\citet{carvalho}), while observing that such centroids exhibit better interpretability by having fewer parameters among the alternative models.  

Working top-down, we demonstrate and validate on several datasets a novel approach to summarizing ensembles of GLMs.  Our data shows this approach can result in models nearly identical to full ensembles in performance and vastly superior in interpretability, owing to dramatically reduced ensemble sizes (Figure \ref{Figure2}).  In addition, since this approach can operate on any models that can be shoehorned into a Cartesian space, it shows promise for compressing and thus summarizing ensembles of other types - for instance, causal model ensembles with individual models represented as orderings (\citet{teyssier}).  We posit that our approach can alter applicability of ensemble methods in general, making their use possible for a wide range of applications where the bottleneck has been the interpretability of results. 

Future directions of research may include multivariate classification methods beyond GMM and hierarchical clustering, as well as extension of this methodology beyond ensembles of GLMs to other types of predictive ensembles.

\section{Comments}
Presented at NIPS 2016 Workshop on Interpretable Machine Learning in Complex Systems

\section{Tables and Figures}
\begin{table}[ht]
\centering
\begin{tabular}{rllllll}
  \hline
 & adenocarcinoma & breast.2.class & colon & leukemia & prostate & P-value \\ 
  \hline
randomGLM & 0.67 & 0.7 & 0.89 & 1 & 0.95 & - \\ 
  medoid-compressed & 0.63 & 0.64 & 0.86 & 0.98 & 0.92 & 0.004 \\ 
  centroid-compressed & 0.67 & 0.67 & 0.79 & 0.99 & 0.92 & 0.07 \\ 
   \hline
\end{tabular}
\caption{AUCs by method and dataset.  Last column shows p-value of one-tailed paired t-test vs randomGLM ($H_0$: the compressed ensemble performs as well as the full ensemble).  Values are averaged over 3 folds and 3 repeats.  Because of multiple repeats, standard errors are not shown.} 
\label{table:Table1}
\end{table}

\newcommand\cropped[1]{%
    \immediate\write18{convert -trim #1.pdf #1cropped.pdf}%
    \includegraphics[width=0.7\linewidth]{#1cropped.pdf}}
    
\begin{figure}[h]
  \fbox{\cropped{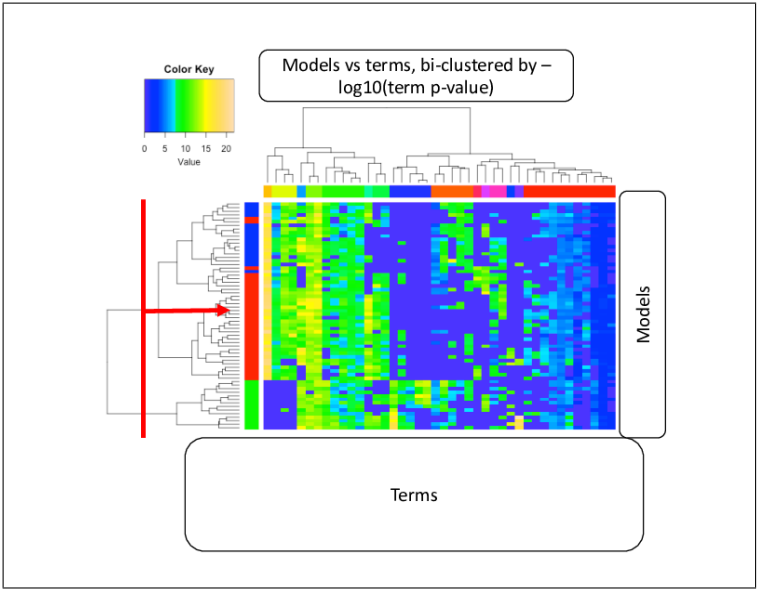}}
 \caption{Fitting model clusters by traversal of the hierarchical clustering tree.  Each cut through the dendrogram implies a column vector of cluster membership that can be used to predict significances associated with the model coefficients of the GLM ensemble (heatmap).}
 \label{Figure1}
\end{figure}

\begin{figure}[h]
\fbox{\cropped{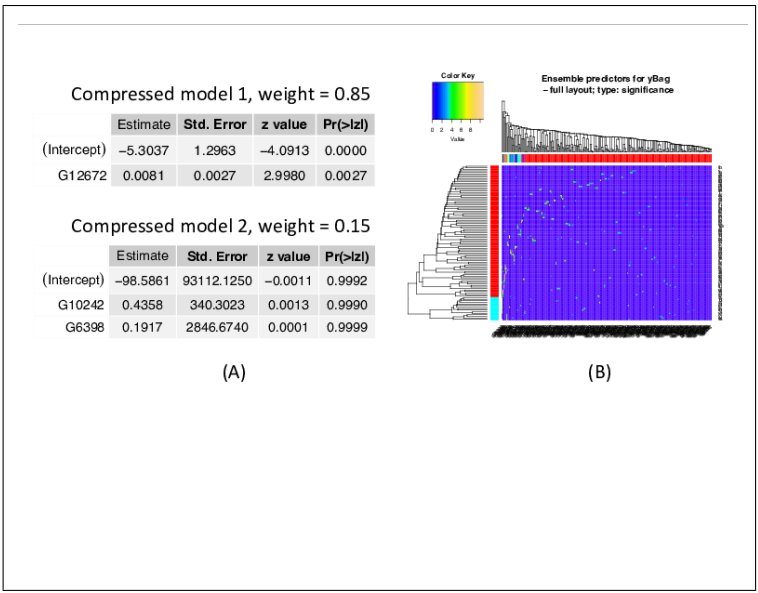}}
 \caption{Benefits of compression (adenocarcinoma dataset): (A) compressed ensemble represented by cluster medoids has only three predictors across just two models and (B) uncompressed ensemble has an uniterpretably large number of predictors, most rare.  The two ensembles have mean OOS AUCs of 0.63 and 0.67, respectively.  The centroid-represented ensemble (not shown) is tied with the uncompressed one OOS (see Table \ref{table:Table1}).  Note that the compressed ensemble, being easy to inspect, illustrates overfitting occurring in both in the uncompressed and the compressed ensembles, as evidenced by very high standard errors on term coefficients in network 2 of the compressed ensemble.  This is probably due to a combination of not sufficiently robust methods (randomGLM uses AIC-guided stepwise search) and p >> n with a very small n.  Compressed ensemble's digestible size allows for thorough manual inspection.}
 \label{Figure2}
\end{figure}

\small

\clearpage

\section*{Acknowledgements}

The authors would like to acknowledge Leon Furchtgott and Fred Gruber for their invaluable feedback on the manuscript, and Fred Gruber for his help with \LaTeX.

\bibliographystyle{plainnat}

\end{document}